\documentclass{article}

\usepackage{PRIMEarxiv}

\usepackage[utf8]{inputenc} 
\usepackage[T1]{fontenc}    
\usepackage{hyperref}       
\usepackage{url}            
\usepackage{booktabs}       
\usepackage{amsfonts}       
\usepackage{nicefrac}       
\usepackage{microtype}      
\usepackage{fancyhdr}       
\usepackage{graphicx}       
\graphicspath{{media/}}     

\title{Effect of temporal resolution on the reproduction of chaotic dynamics via reservoir computing
}

\author{
  Kohei Tsuchiyama, Andr\'{e} R\"{o}hm,Takatomo Mihana, Ryoichi Horisaki, Makoto Naruse \\
  Department of Information Physics and Computing, Graduate School of Information Science and Technology\\
  The University of Tokyo, 7-3-1 Hongo, Bunkyo-ku\\
  Tokyo\\
  \texttt{tsuchiyama-kohei655@g.ecc.u-tokyo.ac.jp} \\
}

\begin{document}
\maketitle
\begin{abstract}
    Reservoir computing is a machine learning paradigm that uses a structure called a reservoir, which has nonlinearities and short-term memory. 
    In recent years, reservoir computing has expanded to new functions such as the autonomous generation of chaotic time series, as well as time series prediction and classification. 
    Furthermore, novel possibilities have been demonstrated, such as inferring the existence of previously unseen attractors. 
    Sampling, in contrast, has a strong influence on such functions. 
    Sampling is indispensable in a physical reservoir computer that uses an existing physical system as a reservoir because the use of an external digital system for the data input is usually inevitable. 
    This study analyzes the effect of sampling on the ability of reservoir computing to autonomously regenerate chaotic time series. 
    We found, as expected, that excessively coarse sampling degrades the system performance, but also that excessively dense sampling is unsuitable. 
    Based on quantitative indicators that capture the local and global characteristics of attractors, we identify a suitable window of the sampling frequency and discuss its underlying mechanisms.
\end{abstract}


\section{\label{sec:level1}Introduction}
Reservoir computing\cite{Jaeger2010,Maass2002} has been intensively studied because of its fast-learning ability and simple architecture.
Reservoir computing is typically composed of three layers: input, reservoir, and output.
One of the simplest examples is the so-called echo state network (ESN). 
Here, the reservoir layer consists of a massive number of randomly-connected nonlinear nodes.
Because the reservoir provides a recursive structure, temporal memorization and mixing of incoming data may occur.
During training of the reservoir computing system, only the weights of the output matrix are reconfigured, whereas the weights in the reservoir and input layer are untouched, leading to simple learning rules. 
The properties of randomly-connected ESNs with simple learning principles are compatible with physical implementations of equivalent processes in a wide variety of substrates ranging from electrons, to fluids, photons\cite{Sunada2021} and more. 

Reservoir computing can be used for prediction in chaotic dynamical systems.\cite{Nakajima2021}  
While chaotic dynamical systems, such as weather and arrhythmia, are impossible to predict exactly over long time scales,\cite{Tanaka2021} short-term and qualitative predictions of chaotic systems to a certain degree are considered useful. 
For example, weather prediction has attracted attention\cite{Jaurigue2022} owing to the vast financial impact that extreme weather events can have.
In the early days of reservoir computing, Jaeger and Haas revealed that reservoir computing can reproduce a chaotic time series over a short time period,\cite{Jaeger2004} and many studies have improved upon their results.\cite{Tanaka2021,Haluszczynski2019,Griffith2019}

In dynamical systems, there is the notion of attractors, which are the long-term patterns where most trajectories travel over time in the phase space of the system.
Reservoir computing can reproduce attractors\cite{Lu2018} with high accuracy.\cite{Gonon2020} 
Attractor reproductions are also important for the analysis and control of chaotic systems.\cite{Abarbanel1993} 
One point of interest here is the high similarity of the original and reproduced attractors. 
Remarkably, a properly trained reservoir computer can succeed in predicting the existence even of attractors that were not used in the training phase.\cite{Rohm2021}  
Therefore, previously unseen attractors can be autonomously generated by reservoir computing. 

As mentioned, the physical realization of reservoir computing is of high interest for exploiting the performance of these systems, including optoelectric systems,\cite{Larger2012} 
soft materials,\cite{Nakajima2015,Kawase2021} 
spin-based systems,\cite{Nakane2018} and optical processes.\cite{Rafayelyan2020,Cunillera2019}
The applications of reservoir computing are diverse, but time series prediction is a critical application because the memorization properties of the reservoir are exploited in this process. 

When trying to reproduce a system in reservoir computing, the temporal sampling of the original data is one of the most fundamental aspects. 
For example, one may assume that the sampling frequency should always be as high as possible to maximize information. 
However, a large number of points would then be required to cover all the relevant dynamical time scales. 
Therefore, some inherent trade-offs are made when considering the appropriate sampling frequency. 
In this study, we examine the effect of sampling in reservoir computing on the quality of the autonomous reproduction of chaotic time series and propose a method for designing the proper sampling frequency.


The remainder of the paper is organized as follows. 
First, we introduce methods to reproduce chaotic attractors using reservoir computing in Sec. \ref{section:theory}.
Quantitative measures used to evaluate the short-term and attractor structure reproduction quality are proposed in Sec. \ref{section:indicator}.
We then demonstrate that the reproduction ability of this system changes for different sampling frequencies in Sec. \ref{section:numerical experiment}.
Finally, we discuss the results and propose a method for setting the sampling frequency to obtain favorable reservoir computing performance in Sec. \ref{section:discussion}. 


\section{Autonomous reproduction of chaotic time series}
\label{section:theory}
\subsection{Chaotic systems}
\begin{figure}
\centering
\includegraphics[width=9cm]{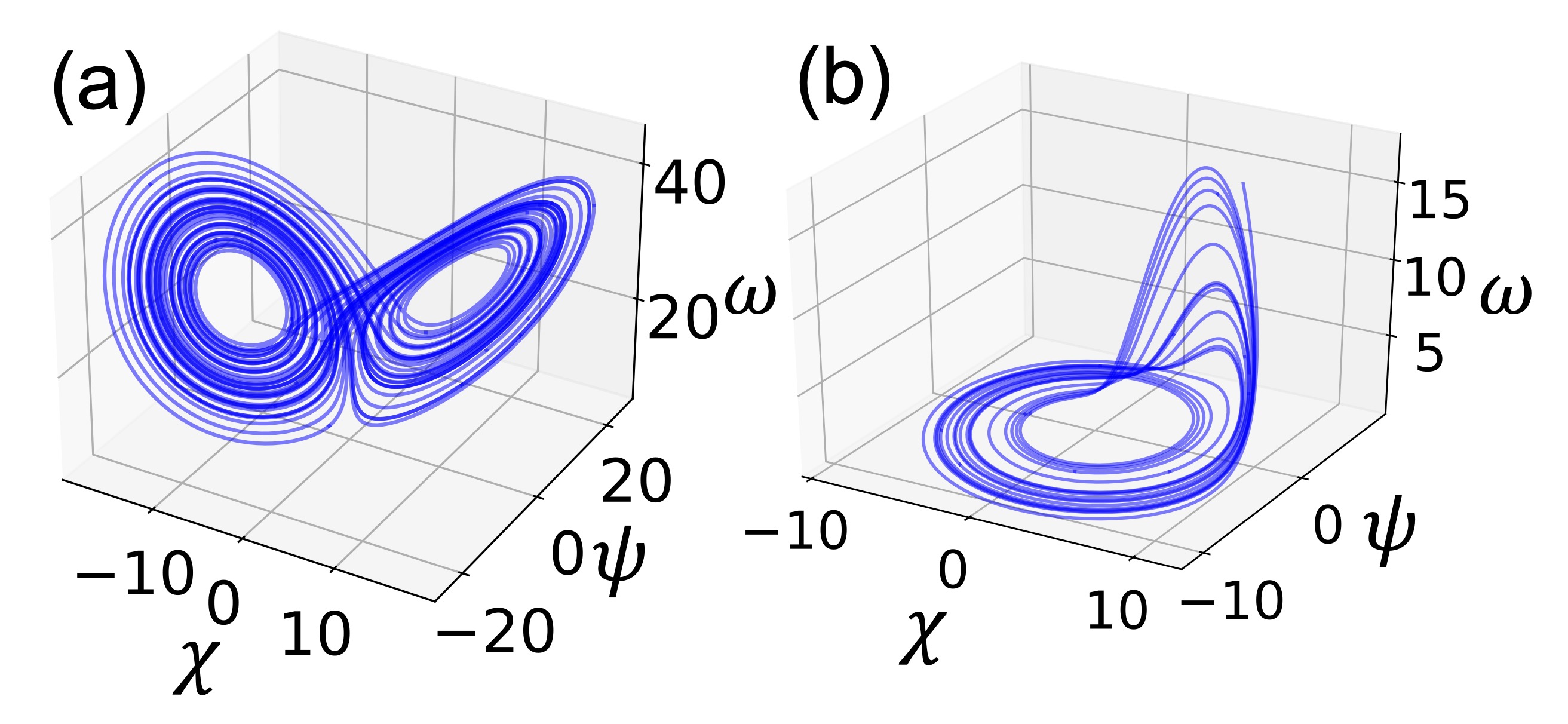}
\caption{\label{fig:attractor} 
Example of a time series using the strange attractor of (a) the Lorenz attractor using Eqs.~(\ref{eq:Lorenz-x})-(\ref{eq:Lorenz-z}), and (b) the R\"{o}ssler attractor using Eqs.~(\ref{eq:Rossler-x})-(\ref{eq:Rossler-z}).
}
\end{figure}

Chaotic systems are a special class of nonlinear dynamical systems characterized by exhibiting a large dependence of the time series on minute differences in the initial state.
This idea of chaos is also connected to the notion of strange attractors. 
In this study, the Lorenz\cite{Lorenz1962} and R\"{o}ssler\cite{Rossler1976} system are used as examples of chaotic dynamical systems.
Figure~\ref{fig:attractor} shows the two strange attractors thereof.

The Lorenz system is a set of three coupled ordinary differential equations (ODEs) given by 
\begin{equation}
  \frac{d\chi}{dt} = -p\chi + p\psi \label{eq:Lorenz-x}
\end{equation}
\begin{equation}
  \frac{d\psi}{dt} = -\chi\omega + r\chi -\psi \label{eq:Lorenz-y}
\end{equation}
\begin{equation}
  \frac{d\omega}{dt} = \chi\psi - b\omega. \label{eq:Lorenz-z}
\end{equation}
The corresponding Lorenz attractor is shown in Fig.~\ref{fig:attractor}(a), exhibiting the famous butterfly-like shape.
We use the standard parameters $p = 10, r = 28,$ and $b = 8/3$.

Conversely, the R\"{o}ssler system is defined as
\begin{equation}
  \frac{d\chi}{dt} = -\psi - \omega   \label{eq:Rossler-x}
\end{equation}
\begin{equation}
  \frac{d\psi}{dt} = \chi + a\psi   \label{eq:Rossler-y}
\end{equation}
\begin{equation}
  \frac{d\omega}{dt} = b + \chi\omega - c\omega.  \label{eq:Rossler-z}
\end{equation}
The R\"{o}ssler attractor is shown in Fig.~\ref{fig:attractor}(b).
We set the parameters as $a = 0.2, b = 0.2,$ and $c = 5.7$.

If the state is expressed as $r=(\chi,\psi,\omega)$, we can rewrite these equations as $\frac{d}{dt}r=f(r)$.
The function $f$ corresponds to either of the above dynamical systems.

The chaotic motions of both systems are characterized by certain irregular oscillations of the state variables. 
The time scales of these oscillations are indirectly defined by these equations and the parameters used.

\subsection{Reservoir computing model}
\begin{figure}
\centering
\includegraphics[width=9cm]{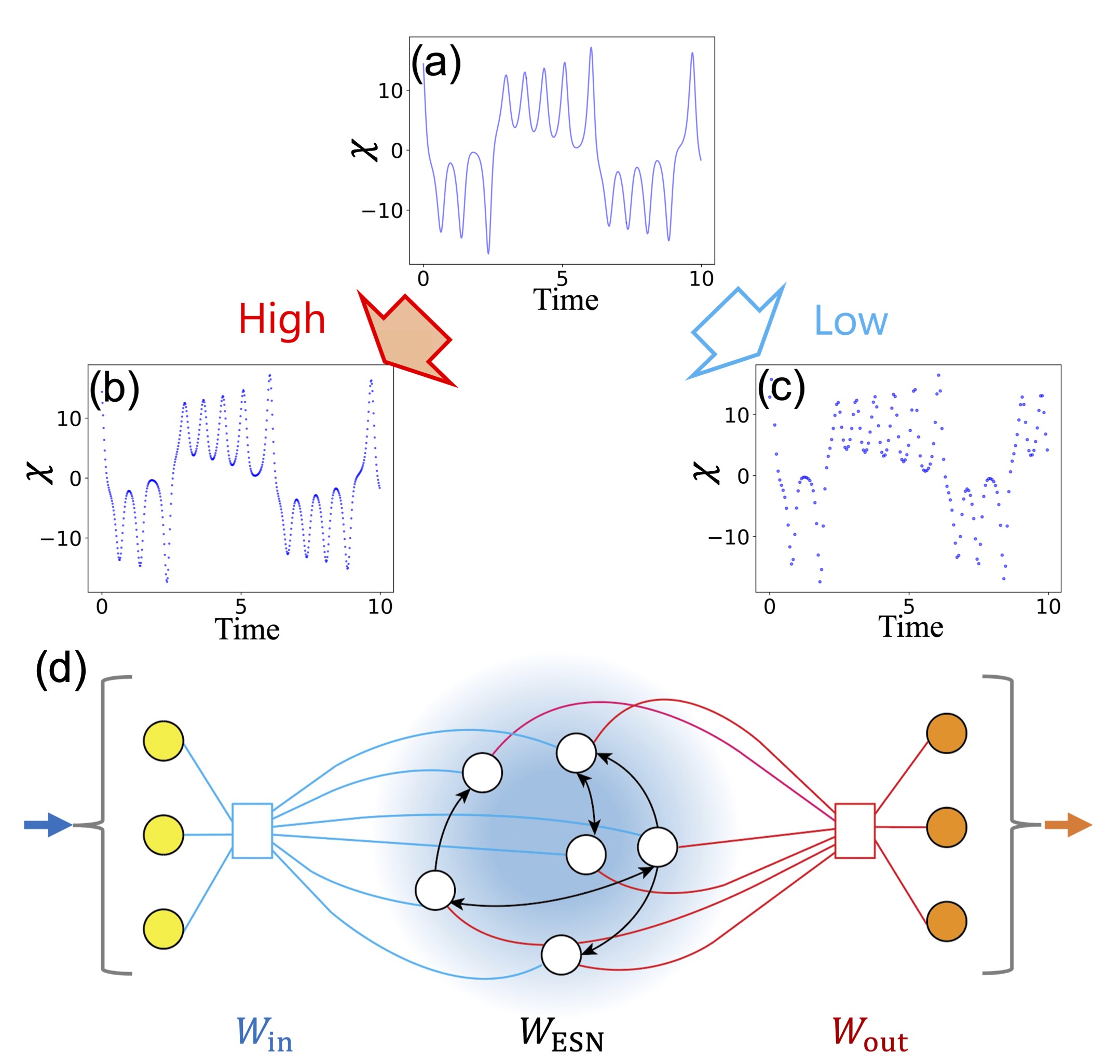}
\caption{\label{fig:reservoir} 
(a,b,c) Schematic illustration of the sampling of a continuous time series. 
(a) Continuous time series. 
(b,c) Examples of sampled time series with (b) a relatively short sampling interval and (c) a relatively large sampling interval.  
(d) Architecture of reservoir computing. 
}
\end{figure}

Figure~\ref{fig:reservoir}(d) shows a schematic illustration of the architecture of the reservoir computing system.
Here, we assume that the reservoir consists of an ESN of $N\in\mathbb{N}$ randomly-connected nonlinear nodes.
Each node has a real-valued state variable $x_i\in\mathbb{R}(i=1,2,...,N)$, and these $N$ nodes are connected in a sparse manner.
The state of the reservoir can be expressed as an $N$-dimensional vector $X=[x_1,x_2,x_3,...,x_N]^{\top}$ where $\top$ means transpose.
These states evolve with a nonlinear function.
In the present study, we use the hyperbolic tangent function as the nonlinearity in the system.
Specifically, the state of the reservoir is updated as
\begin{equation}
  X[n+1] = \tanh{(W_{\mathrm{ESN}} X[n] + G\cdot W_{\mathrm{in}} u[n] + \mathbf{b})}\in\mathrm{R}^N\label{eq:X_state_evolve}
\end{equation}
where $n$ is a time step, and $W_{\mathrm{ESN}}$ is an $N\times N$ sparse matrix expressing the weights of connections among nodes.
In this study, $2\%$ non-zero elements are used, and it is normalized
such that the absolute value of the largest eigenvalue is $0.95$. 
The term $G$ is a scalar gain factor for the inputs, which is set to $0.2$ in this study; $W_{\mathrm{in}}$ is an $N\times U$ matrix determining the input weights, whose elements are expressed by $W_{\mathrm{in}}^{ij}$, where $i\in\{1,2,...,N\}$ and $j\in\{1,2,...,U\}$; $u[n]$ is the $U$-dimensional input at the $n$-th step; and $\mathbf{b}=[b_1,b_2,...,b_N]$ is an $N$-dimensional bias vector. We choose $W_{\mathrm{in}}^{ij}\sim U[-1,1]$ and $b_i\sim U[-0.3,0.3]$.

In addition, the output is calculated as
\begin{equation}
  y[n] = X[n]^{\top}W_{\mathrm{out}},
\end{equation}
where $y[n]$ is the $M$-dimensional output for step $n$,
$W_{\mathrm{out}}$ is an $N\times M$ matrix used to determine the output values, and the elements therein are determined in the supervision or learning phase.
In the following, $N, G$ are regarded as reservoir parameters, while $W_{\mathrm{ESN}}, W_{\mathrm{in}}, \mathbf{b}$ are determined randomly with these parameters.

\subsection{Learning in reservoir computing}
As stated above, the reservoir computer learns to predict the dynamical system by fitting the weight of the output matrix $W_{\mathrm{out}}$.
To achieve this, it is necessary to consider the dynamical change of the reservoir state.

The state matrix of the reservoir $S\in \mathbb{R}^{K\times N}$ is defined as
\begin{equation}
  S = \left[ X[1], X[2],\dots,X[K] \right]^{\top},
\end{equation}
where $K$ is the length of the input time series.
That is, this state matrix encodes the response of the reservoir to the input time series.
A predicted time series is expressed as a $K$-dimensional vector.
Using $S$ and $W_{\mathrm{out}}$, the output time series $Y\in \mathbb{R}^{K\times M}$ is expressed as
\begin{equation}
Y = SW_{\mathrm{out}}.
\end{equation}
At this point, it is possible to compare the reservoir system output $Y$ with the original system time series.
It is then possible to determine $W_{\mathrm{out}}$.
In this study, we use ridge regression (see Appendix \ref{Appendix:riddge}).
Once $W_{\mathrm{out}}$ has been determined, the reservoir computing system has finished training.

\subsection{Sampling interval}
As an example, consider the target output of the reservoir to be a chaotic time series of a Lorenz system.
The state variables $r(t)$ of the physical system are in principle time-continuous (Fig.~\ref{fig:reservoir}(b)). 
However, this time-continuous raw output cannot be used as target data.
First, we cannot actually calculate the time series from the ODEs as a time-continuous object.
All information needs to be discretized for processing in digital computers.
Sampling in the time-domain is the process of obtaining a discrete sequence of points from a continuous time series.
The time resolution of the sampling process is quantified by the length of time between two successive samples, which is referred to hereafter as the sampling interval (SI).  
Figures~\ref{fig:reservoir}(c) and (d) illustrate  high- and low-frequency sampling or a small and large SI, respectively.

In practice, all used time series should also have a buffering component ($K_{\mathrm{buffer}}$ steps) because the reservoir needs time to begin storing information of the input data.
If the training succeeds, the reservoir computer can then make next-step predictions with a low error.
With this learned $W_{\mathrm{out}}$, the reservoir computer can then also be used in the autonomous mode.

\subsection{Autonomous mode}
\begin{figure}
\centering
\includegraphics[width=9cm]{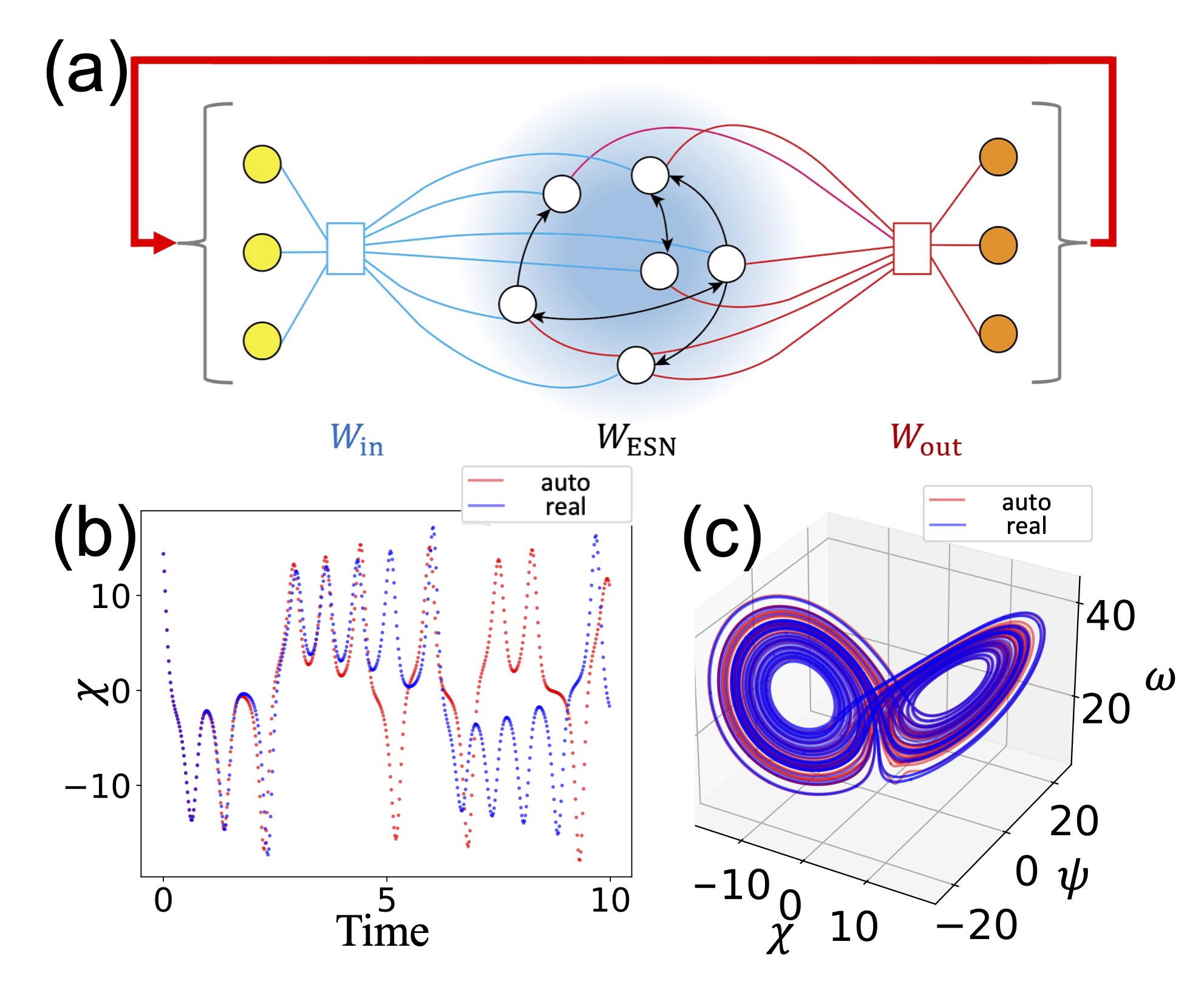}
\caption{\label{fig:reproduction}
Autonomous reproduction of a chaotic time series. 
(a) Schematic illustration of the autonomous mode operation of reservoir computing. 
The output of each time step is returned into the system as the next input.
(b) Time evolutions of the Lorenz system's $\chi$ element generated by the original Eqs.~(\ref{eq:Lorenz-x})-(\ref{eq:Lorenz-z}) (blue) and regenerated by autonomous mode reservoir computing (red). 
In the short term, both values are similar to each other (short-term prediction), but they exhibit different traces in the long term. 
(c) Blue and red curves indicate the attractors generated by the original chaos dynamics and the autonomous mode of reservoir computing, respectively. They exhibit similar shapes (long-term structural prediction).  
The sampling interval used for the reservoir is $SI = 0.005$.
}
\end{figure}

The autonomous mode of reservoir computing is the case in which the output of the reservoir acts as the input of the same reservoir, which is represented by the red arrow in Fig.~\ref{fig:reproduction}(a). 
If the reservoir was successfully trained so that given a state $r(t)$ of a dynamical system, it predicts the next step $r(t + SI)$ with high accuracy, then the autonomous mode of that reservoir can then reproduce the dynamical system without input. 
Therefore, a reservoir that learned to predict the next time step of the Lorenz system, for example, can reproduce the shape of the Lorenz attractor in the autonomous mode. 
However, owing to the basic properties of chaotic systems, the long-term trajectories are not identical to the genuine time series of the Lorenz dynamical system, even when starting from the same initial conditions.

The blue curve in Fig.~\ref{fig:reproduction}(b), denoted as ``real,'' shows the Lorenz attractor computed by the genuine Lorenz equations.
In the training phase, the reservoir is trained such that the output predicts one step ahead of the Lorenz system. 
The red curve in Fig.~\ref{fig:reproduction}(b), denoted as ``auto,'' represents the attractor reproduced by the autonomous mode of reservoir computing, exhibiting a similar trace to the genuine attractor denoted by the blue curve. 

Figure~\ref{fig:reproduction}(b) shows the time evolution of the $\chi$-value of the Lorenz system, where the original (``real'') and autonomous mode (``autonomous'') agree well from the initial time until approximately $t=4$. Thereafter, these two trajectories exhibit quite different values; however, as we observe in Fig.~\ref{fig:reproduction}(c), it should be noted that the butterfly shape of the attractor itself is well recovered by the autonomous mode. 

In summary, the autonomous-mode reservoir system can closely reproduce the original time series in a short-term regime, which is referred to as short-term prediction.
In addition, whereas accurate long-term prediction precision is difficult to achieve, the autonomous-mode reservoir system can yield a similarly shaped attractor in the long term as demonstrated in Fig.~\ref{fig:reproduction}(b), which shows a type of long-term structural prediction.

\section{Indicators for the attractor reproduction ability}
\label{section:indicator}

\subsection{Short-term reproduction}
The mean prediction horizon (MPH) is a measure of how long the reservoir accurately reproduces the chaotic system under study once it is executed in the autonomous mode when initialized from and compared to a sample time series.\cite{Wyffels2010}
The time-dependent error between the predicted and original chaos time series is calculated by the normalized mean squared error (NMSE) defined by
\begin{equation}
  NMSE(t) 
  = \frac{\sum_{i=1}^{N_{\mathrm{trial}}}(y_{i}^{\mathrm{real}}(t)-y_{i}^{\mathrm{auto}}(t))^2}
          {\sum_{i=1}^{N_{\mathrm{trial}}}\left(y_{i}^{\mathrm{real}}(t)-\overline{y_{i}^{\mathrm{real}}(t)}\right)^2},
\end{equation}
where 
$N_{\mathrm{trial}}$ is the total number of time series used, 
$y_{i}^{\mathrm{real}}(t)$ 
is the genuine or real signal level of the dynamical system under study at time $t$ of the $i$-th trial, 
$y_{i}^{\mathrm{auto}}(t)$ denotes the signal level generated by the autonomous-mode reservoir computing system at time $t$ of the $i$-th trial, and $\overline{y_{i}^{\mathrm{real}}(t)}$ is the mean of $y_{i}^{\mathrm{real}}(t)$ over the $N_{\mathrm{trial}}$ time series. 
The smaller the NMSE, the more accurately the reservoir computer reproduces the genuine dynamical system. 

The MPH then quantifies how long the accurate prediction is accomplished once the system is operated in the autonomous mode.
In other words, the MPH is defined as the length of time for which the NMSE stays consecutively below a presupposed threshold $h$.
\begin{equation}
  MPH(h) = \mathrm{min}\ t \ \ s.t.\ NMSE(t)>h. \label{eq:MPH}\nonumber
\end{equation}
We use $h = 0.5$ for the Lorenz or $h = 0.1$ for the R\"ossler system, as displayed in Table~\ref{tab:table1}.
By this definition, $MPH$ shows the ability of the system to achieve short-term reproduction.

\subsection{Long-term reproduction}
In the long term, reservoir computing can reproduce the shape of a chaotic attractor, although the generated time series differs from the original one. 
Here, we examine the structure of the attractor from two perspectives: the global shape and local structure.

To examine the global shape of the attractor, we introduce the idea of chaotic attractor density.
The chaotic systems in the present study are expressed in $\mathbb{R}^3$ space.
Suppose that there are in total $L$ points in the space. 
Now we divide the space into an array of tiny volumes. 
When there are $M_l$ points in the area denoted by $l$,
the density of points in the area $l$ is defined by
\begin{equation}
    \rho_{l} = \frac{\frac{M_l}{L} + \epsilon}{\sum_l ( \frac{M_l}{L}+\epsilon)}.
    \label{eq:density}
\end{equation}
Herein, $\epsilon$ is a small number and it is introduced to avoid division by zero.
In this study, $\epsilon$ is fixed as $10^{-8}$, which is much smaller than 
$1/L=1/20000=0.00005$.
In this definition, the sum over all volumes $l$ of the density $\rho_l$ is almost $1$ when $\epsilon\approx 0$.
Therefore, $\rho$ can also be interpreted as the probability distribution of where to find the state of the system at a random point in time, assuming stationarity.

\subsubsection{L1 norm}
One way to quantify the difference between two attractors is to use the L1 norm.
Based on the density of a chaotic attractor defined by Eq. (\ref{eq:density}), the L1 norm is given by: 
\begin{equation}
    L1 = \sum_l |\rho_l^{\mathrm{auto}} - \rho_l^{\mathrm{real}}|
\end{equation}
where $\rho_l^{\mathrm{real}}$ indicates the density of the attractors generated by the original chaotic dynamics whereas $\rho_l^{\mathrm{auto}}$ is that regenerated by the autonomous mode of the reservoir computing system. 

\subsubsection{Kullback--Leibler divergence}
As discussed above, the distribution of the attractor density, defined by Eq. (\ref{eq:density}), can be considered as a probability distribution.
The Kullback--Leibler divergence (KLD) is a way of quantifying the distances between probability distributions.\cite{InformationTheory}  
In this study, two kinds of KLDs are examined, which are defined as:
\begin{equation}
    KLD(\mathrm{real}||\mathrm{auto}) = \sum_k \rho_l^{\mathrm{real}}\log{\left(\frac{\rho_l^{\mathrm{real}}}{\rho_l^{\mathrm{auto}}}\right)}      \label{eq:KLD1}
\end{equation}
\begin{equation}
    KLD(\mathrm{auto}||\mathrm{real}) = \sum_l \rho_l^{\mathrm{auto}}\log{\left(\frac{\rho_l^{auto}}{\rho_l^{real}}\right)}, 
    \label{eq:KLD2}
\end{equation}
These equations capture the distances between the attractor generated by the genuine chaotic dynamics and the attractor reproduced by the autonomous mode of the reservoir computing system. 
The value of $\epsilon$ has some effect on KLD.
However, there is no effect when the $\rho$ variables have the same values and the effect is negligible when the ratio 
$M_{k} / M_{\mathrm{total}}$ is large enough in an area $k$.

\subsubsection{Inner product}
The indicators discussed above, namely the L1 norm and KLD, are measures of the difference in the global structures of the attractors. 
We now focus on the local characteristics of the chaotic attractors.

At every step, an inner product is calculated between the normalized local gradient vector obtained by the autonomous-mode reservoir computing system and the normalized gradient vector based on the original chaotic dynamics.
Let the $j$-th point of the prediction by the autonomous reservoir computing system be expressed as $\mathbf{y}[j]$. 
The local gradient calculated by the autonomous reservoir is evaluated by the normalized difference vector, which is defined by $\left(\mathbf{y}[j+1]-\mathbf{y}[j]\right) / \left( \| \mathbf{y}[j+1]-\mathbf{y}[j] \| \right)$.

The gradient vector of the genuine chaotic dynamics at the point $\mathbf{y}[j]$ is computed by $\mathbf{f}(\mathbf{y}[j])$ where the function $\mathbf{f}$ is defined by the right-hand side of the ODEs, such as in Eqs. (1) to (3). 

The indicator IP is defined by the mean of the inner product over time, which is formally given by:
\begin{equation}
    IP = \frac{1}{K}\sum_{j}^{K} \frac{f(\mathbf{y}[j])}{||f(\mathbf{y}[j])||}\cdot \frac{\mathbf{y}[j+1]-\mathbf{y}[j]}{\|\mathbf{y}[j+1]-\mathbf{y}[j] \|}. \label{eq:IP}
\end{equation}
where $K$ is the number of steps.
Based on its definition, $IP$ takes a value in $[-1,1]$; 
When the gradient reproduced by the autonomous mode of the reservoir computing system is parallel to that of the genuine chaotic dynamics everywhere in the attractor, $IP$ is $1$. 
Conversely, if the reproduced gradient is orthogonal to the genuine gradient throughout the time period, $IP$ should be $0$.

\section{SAMPLING INTERVAL-DEPENDENT CHAOTIC DYNAMICS REPRODUCTION}\label{section:numerical experiment}
The Lorenz system with parameters of $p=10$, $r=28$, and $b=8/3$ and the R\"{o}ssler system with $a=0.2$, $b=0.2$, and $c=5.7$ are used as examples for demonstrations in this study.
The number of buffering steps $K_{\mathrm{buffer}}$ is $3000$. 
The length of the time series $K$ is $20000$.
The number of nodes $N$ is $1500$. 
The gain factor $G$ in Eq. (\ref{eq:X_state_evolve}) is $0.2$.
The smallest used SI and the threshold for determining the MPH ($h$) are configured differently when examining the Lorenz and R\"{o}ssler systems, as summarized in Table \ref{tab:table1}.

\begin{table}
\caption{\label{tab:table1}Parameters used in numerical experiments.}
\centering
\begin{tabular}{lll}
Chaos system & \mbox{Minimum SI} & \textit{h} \\
\hline
Lorenz      & $0.005 $ & $0.5$ \\
R\"ossler & $0.01$ & $0.1$ \\
\end{tabular}
\end{table}

\subsection{Short-term reproduction}
The MPH is evaluated for measuring the short-term reproduction ability of each system.
For each of the specified sampling intervals, $1000$ different reservoirs were prepared, each of which was trained from a different chaotic time series. 
Sometimes, the training obviously failed and the resulting attractor reconstruction was very poor.
For example, these systems may converge to far-away and wrongly inferred fixed points or large-scale oscillations. 
Due to the nature of the definitions of the MPH, L1 norm, and KLD, these outliers can have an outsized effect on our quantifiers.
Therefore, we introduce a criterion for ignoring those catastrophic failures.
The trial score (TS) is defined as follows:
\begin{equation}
    TS_i\equiv\sum_t NMSE_{\chi}(t)+NMSE_{\psi}(t)+NMSE_{\omega}(t).
\end{equation}
This value is defined for each individual autonomous time series $i$.
We discard trials whose $TS$ value is above 10 times the median $TS$.

\begin{figure}
\centering
\includegraphics[width=9.5cm]{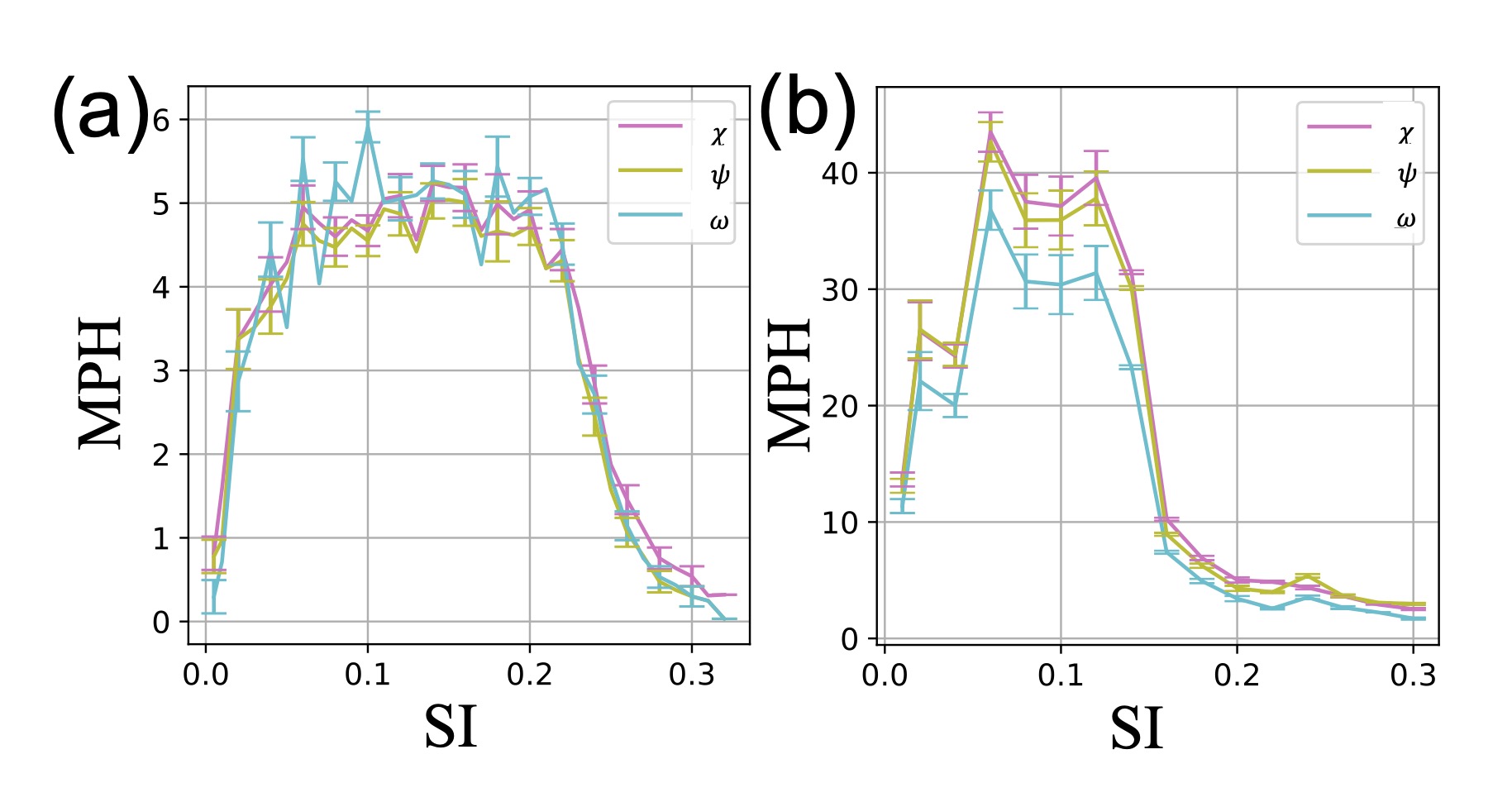}
\caption{\label{fig:MPH}
Mean prediction horizon (MPH) as a function of the sampling interval for evaluating the short-term reproduction ability. (a) Lorenz system. (b) R\"{o}ssler system.}
\end{figure}

We divide those trials into ten data sets.
$MPH$ is calculated for each data set.
In this study, the error bars can be obtained as standard deviations of these ten values.

The red, yellow, and blue curves in Fig.~\ref{fig:MPH}(a) show the MPH as a function of the sampling interval with respect to the $\chi$, $\psi$, and $\omega$ components of the Lorenz system, respectively. 
The $\psi$-axis MPH is measured in the same time units and therefore directly indicates the time length of successful prediction by the reservoir computer.
When compared to Fig.~\ref{fig:reproduction}(b), the maximum MPH of the Lorenz system seen in Fig.~\ref{fig:MPH}(a) exhibits values that exceed the time scale of the chaotic oscillations.
In the same manner, Fig.~\ref{fig:MPH}(b) summarizes the results of the R\"{o}ssler system.
The MPH values for the R\"ossler system are larger in comparison, as it has slower time scales.
When the sampling interval is too big, the MPH exhibits small values. This phenomenon was observed in the case of the Lorenz system with SI greater than $~0.2$ and in the case of the R\"{o}ssler with SI of approximately $0.15$. 

Interestingly, the MPH decreases when the sampling interval is too small. 
For the Lorenz system, the MPH decreases as the sampling interval decreases when the sampling is smaller than approximately $0.1$. 
Likewise, the R\"{o}ssler system shows a similar trend when the sampling interval is less than approximately $0.05$.

All three state variables $\chi$, $\psi$, and $\omega$ show a generally similar shape. 
This supports the assumption that the decrease and increase of the mean prediction horizon are linked and caused by common factors.
This is a natural consequence of the fact that the dynamical equations are coupled.
For stiff systems of equations with large time-scale separations, it may be plausible that the MPH for different variables can have strongly differ, howver this is not the case for the Lorenz nor R\"ossler system.

\subsection{Long-term reproduction}
\begin{figure}
\centering
\includegraphics[width=9cm]{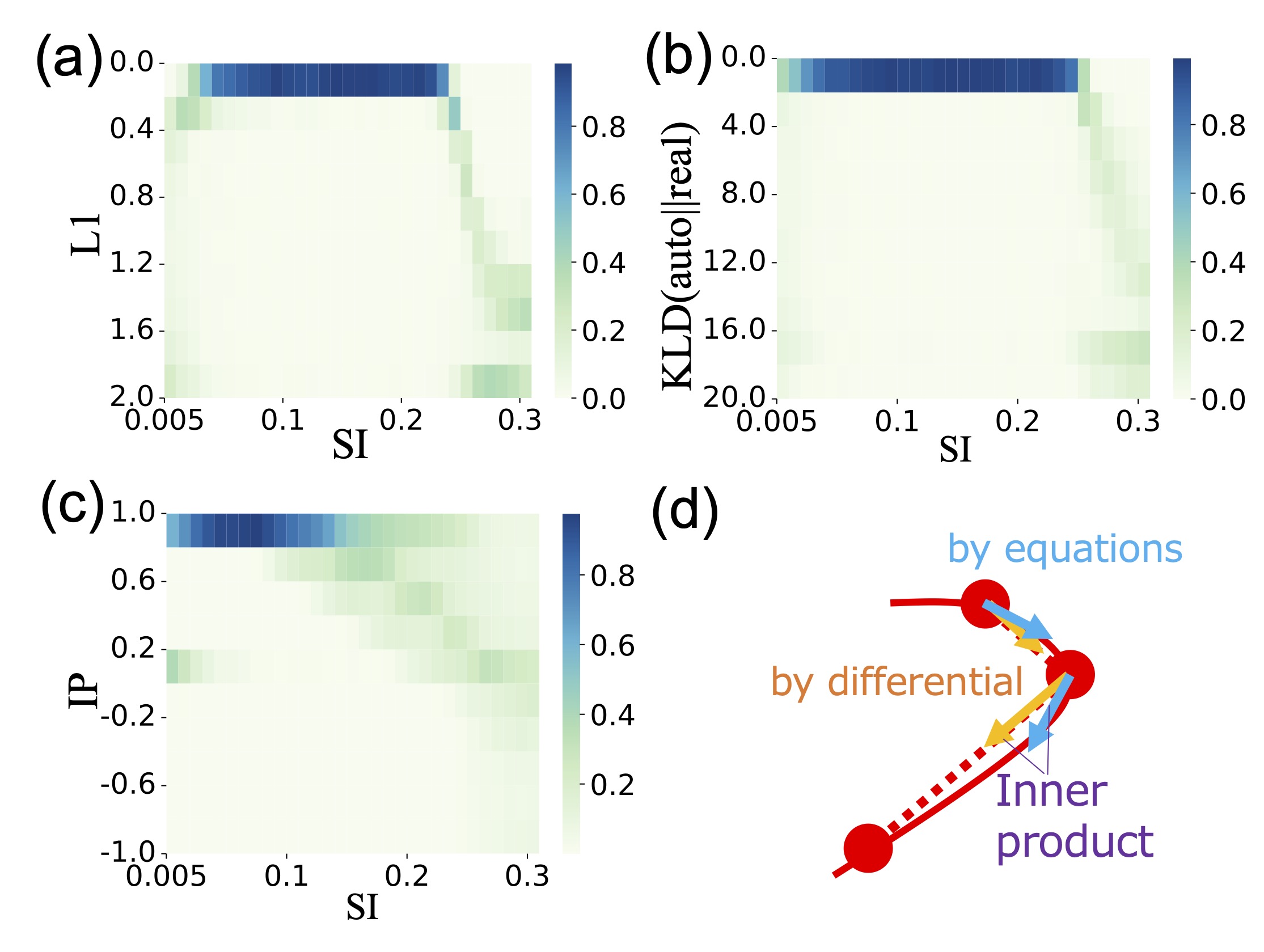}
\caption{\label{fig:D_I} 
Long-term reproduction abilities evaluated by three indicators as a function of the sampling interval. 
(a) L1 norm. 
(b) Kullback--Leibler divergence $KLD(\mathrm{auto}||\mathrm{real})$.
(c) Inner product. 
(d) Schematic illustration of the definition of the inner product of the directions considered in the present study to quantify the local similarities between the attractors regenerated by the autonomous mode reservoir computing system and the genuine chaotic dynamics. 
}
\end{figure}

Here, we present the long-term reproduction abilities of reservoir computing based on the L1 norm and KLD when the chaotic dynamics were the Lorenz system.
Figure~\ref{fig:D_I}(a) shows the incidence heat map of the L1 norm as a function of the sampling interval, with darker blue showing the concentrations of the majority of points.
A value close to $0$ indicates that the probability densities $\rho_l^{\mathrm{real}}$ and $\rho_l^{\mathrm{auto}}$ are highly similar, i.e. a strong topological agreement between the reproduced and original attractor.
A smaller value of the L1 norm is observed when the sampling interval is between approximately $0.005$ and $0.22$.
The autonomous mode reservoir computing system successfully reproduces the attractor when the sampling interval is in this range. 
Conversely, the L1 norm exhibits a large value when the sampling interval is too small or too large. 
Such a trend is similar to that observed in MPH in Fig.~\ref{fig:MPH}(a).
This corresponds to the same SI boundaries that were seen for obtaining a long MPH. 
In other words, the global structure and the local structure are correctly reproduced either at the same time, or neither.

Figure~\ref{fig:D_I}(b) summarizes the KLD defined by Eq. (\ref{eq:KLD2}). 
Similarly to the L1 norm, the KLD exhibits a large value when the sampling interval is too small or too large. 
Therefore, when the sampling interval is in a moderate regime, the KLD exhibits a small value, meaning that the attractors generated by the autonomous mode reservoir computing system and the original chaotic dynamics are similar to each other.

Figure~\ref{fig:D_I}(c) shows the inner product as a function of the sampling interval. 
The inner product exhibits values of almost $1.0$ when the sampling interval is greater than approximately $0.01$ and smaller than $0.15$. 
The inner product of the value $1.0$ means that the reproduced attractors from the autonomous mode reservoir exhibit similar local properties to the original chaotic dynamics.
While the range of a larger inner product is not exactly equal to that of a small L1 norm or KLD, 
the trend is similar between the inner product, L1 norm, and KLD as too small or too large an SI deteriorates the figure of merit.
One reason that the inner product diverges from the real gradient earlier could be related to the nonlinear properties of reservoir computing.
The reservoir computer is likely able to approximate not just a linear step, like an Euler numerical integrator, but also anticipate higher-order terms, similar to a Runge--Kutta scheme. 
However, we do not have access to the local gradient as predicted by the reservoir computer, and therefore the definition of $IP$ given by Eq.~(\ref{eq:IP}) relies purely on the difference of successive points, which will necessarily provide a crude approximation for larger SI values.

Overall, the long-term reproduction is most succesfull for the same values of SI as for maximizing the MPH. 
Therefore, there seems to be no trade-off between short-term and long-term reproduction abilities, but rather a synergistic relationship. 
This indicates that the reservoir computer either correctly learns to predict the dynamical evolution of the reservoir, or fails. 
Considering that the reservoir computer computer was originally trained on a time-series prediction task, it would seem plausible that the global reproduction capability is a consequence of the local prediction quality. 
Would it be possible to reverse this process, i.e., to construct a short-term prediction by training on reproducing the global structures first?
This seems unlikely.
After all, the probability densities $\rho_l$ no longer contain any information about the sequence of points and have thus lost their temporal ordering. 
There may be no way to recover this information and there may be multiple plausible systems that reproduce the same density.

\section{Discussion and Conclusion}
\label{section:discussion}
We demonstrated the effect of sampling on the autonomous reproduction of chaotic time series by reservoir computing.
We quantitatively evaluated the reproduction performance by focusing on the short- and long-term reproduction abilities. 

The MPH demonstrated that a sampling interval that is too small or too large significantly deteriorates the short-term prediction quality.
The long-term reproduction abilities were analyzed by evaluating the L1 norm between the distribution of the time series generated by the autonomous mode reservoir and that of the genuine chaotic dynamics. Similarly, the KLD was also evaluated. 
Similar to the case of MPH, sampling intervals that are too small or too large resulted in decreased similarity, whereas an appropriate SI resulted in small distances between the generated and real attractors.
Furthermore, the local features of the regenerated attractor were examined by evaluating the indicator IP of the gradient, which also exhibited a similar trend to those observed above.


We consider that there are two underlying mechanisms behind such observations.
When the sampling interval is too large (i.e., too scarce sampling), the reservoir cannot reproduce the chaotic time series.
This occurs because the time series used for learning may not have enough information concerning the fine details of the attractor structures.
Therefore, the reservoir cannot identify the shape of the attractors and cannot reproduce them.
Thus, it is expected that the MPH goes to $0$ when SI goes to $\infty$.

Conversely, when the sampling interval is too small (i.e., too dense sampling), the memory of the reservoir may not be capable of holding information over a long time scale.
In other words, there is too much detailed information, and the system cannot obtain the global information necessary for attractor reconstruction.
Therefore, it is expected that the MPH goes to $0$ when SI goes to $0$ (for a given reservoir with limited memory).

In practice, the upper limit on SI is likely more important than the lower limit. 
If a time series is obtained with too short SI, assuming enough points are given, a corresponding time series with larger SI can always be calculated via down-sampling.
Similarly, the memory length of the reservoir could be increased via a variety of means, such as parameter engineering, network size or even using parallel or cascaded reservoirs.

The above findings indicate that the careful consideration of the sampling frequency is important for physical system implementation involving reservoir computing.
It is necessary to set a suitable value for the sampling rate such that the autonomous mode may be used.
In conclusion, it is necessary to set the sampling frequency considering the upper and lower bounds that are determined by the two mechanisms mentioned above.

\section*{Appendix}




\subsection{Ridge regression}\label{Appendix:riddge}
Let the prediction equation be $\hat{\mathbf{y}}=X\mathbf{w}$ ($\hat{\mathbf{y}}\in\mathbb{R}^{K\times 3},X\in\mathbb{R}^{K\times N},\mathbf{w}\in\mathbb{R}^{N\times 3}$).
In multiple regression analysis, we want to find $\mathbf{w}$ that minimizes the loss function $L$,
defined as
\begin{equation}
L=\sum_{i}(y_i-\hat{y}_i)^2\\
=(\mathbf{y}-X\mathbf{w})^T(\mathbf{y}-X\mathbf{w})\\
=\mathbf{w}^TX^TX\mathbf{w}-2\mathbf{y}^TX\mathbf{w}+\mathbf{y}^T\mathbf{y}.
\end{equation}
Therefore, the derivative of $L$ is 
\begin{equation}
\frac{\partial L}{\partial\mathbf{w}}=2X^TX\mathbf{w}-2X^T\mathbf{y}.
\end{equation}
Because the differential should be zero when the loss $L$ is minimal, it follows that $\mathbf{w}=(X^TX)^{-1}X^T\mathbf{y}$.

With regularization (i.e.,  ridge regression), the loss is slightly modified to $L=\sum_i(y_i-\hat{y}_i)^2+\alpha\mathbf{w}^T\mathbf{w}$.
Then, the optimal weights are $\mathbf{w}=(X^TX+\alpha\mathbf{I})^{-1}X^T\mathbf{y}$, where $\alpha$ is called the regularization factor.
In this study, $\alpha$ is fixed at $0.01$.




\subsection{Fourier Transformation}
\begin{figure}
\centering
\includegraphics[width=8cm]{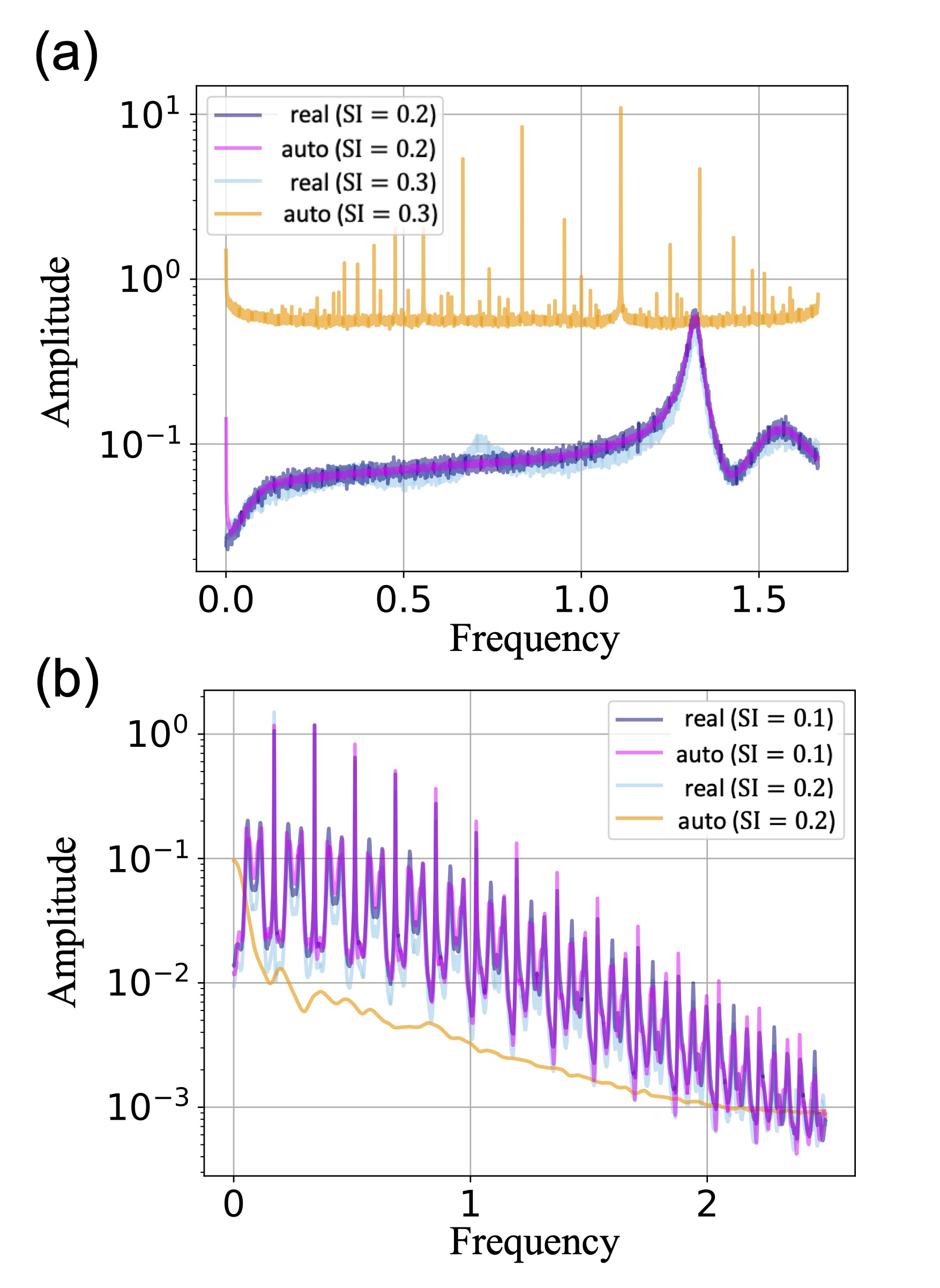}
\caption{\label{fig:fft} 
Mean values of the amplitude spectrum of the time series generated by the autonomous mode reservoir (auto) and original chaotic dynamics (real).
(a) The $\omega$ value amplitude of Lorenz chaos when SI are set as $0.2$ and $0.3$.
(b) The $\omega$ value amplitude of R\"{o}ssler chaos when the SI are set to $0.1$ and $0.2$.
}
\end{figure}


We investigated the reconstruction ability in $\mathbb{R}^3$ space in Sec. \ref{section:numerical experiment}.
There is a strong relationship between the sampling intervals and frequency elements of the dynamics.
We conducted a Fourier analysis of the autonomous mode reservoir computing system in both successful and failed reconstruction cases by averaging over $1000$ trials. 
The magenta curve in Fig.~\ref{fig:fft} shows the amplitude spectrum of the time series generated by the autonomous mode reservoir computing system, whereas the navy curve therein is that of the original chaotic dynamics when SI is in the successful reconstruction range of $L1$ discussion.
Here, Fig.~\ref{fig:fft}(a) shows the results for the Lorenz and Fig.~\ref{fig:fft}(b) for the R\"{o}ssler system, respectively.
On the other hand, the orange curve in Fig.~\ref{fig:fft} shows that of the reproduced attractor, whereas the skyblue curve is that of the original chaotic dynamics when SI is too large.
The Fourier spectra are consistent with the other indicators presented in this paper: They show that frequency elements can be reproduced when the SI is of intermediate size.



\section*{Acknowledgments}
This study was supported in part by the CREST project (JPMJCR17N2) funded by the Japan Science and Technology Agency, Grants-in-Aid for Scientific Research (JP20H00233); and Transformative Research Areas (A) (JP22H05197) funded by the Japan Society for the Promotion of Science. A.R. was funded by the Japan Society for the Promotion of Science as an International Research Fellow.

\bibliographystyle{unsrt}  
\bibliography{references.bib}

\end{document}